\pdfoutput=1
\documentclass[11pt]{article}
\usepackage[final]{acl}

\usepackage{times}
\usepackage{latexsym}
\usepackage{booktabs}
\usepackage{nicematrix}
\usepackage{array}
\usepackage{enumitem}

\usepackage[T1]{fontenc}

\usepackage[utf8]{inputenc}
\usepackage{microtype}
\usepackage{inconsolata}
\usepackage{graphicx}
\usepackage{todonotes}
\usepackage{arydshln} 
\usepackage{pgf-pie}
\usepackage{pifont}
\usepackage{pgfplots}
\pgfplotsset{compat=1.18} 
\usepackage{xurl}
\usepackage{xcolor}
\usepackage{amssymb}
\usepackage{subcaption}

\usepackage{algorithm}
\usepackage{algpseudocode}
\usepackage{amsmath}

\newcommand{\ourkb}{\mbox{\textsc{GPTKB}}}
\newcommand{\website}{\url{https://gptkb.org}}

\definecolor{gg-red}{RGB}{234,67,53}
\definecolor{gg-blue}{RGB}{66,133,244}
\definecolor{gg-yellow}{RGB}{251,188,4}
\definecolor{gg-green}{RGB}{52,168,83}
\definecolor{gg-orange}{RGB}{255,109,1}
\definecolor{gg-teal}{RGB}{70,189,198}

\title{Enabling LLM Knowledge Analysis via Extensive Materialization}

\author{
Yujia Hu$^{1}$
\qquad
Tuan-Phong Nguyen$^2$
\qquad
Shrestha Ghosh$^3$
\qquad
Simon Razniewski$^1$
\\ 
\\
$^1$ScaDS.AI \& TU Dresden, Germany
\\
$^2$Max Planck Institute for Informatics, Saarbr\"ucken, Germany
\\
$^3$ University of Tübingen, Germany
\\
{\small\texttt{\{yujia.hu,simon.razniewski\}@tu-dresden.de \quad tuanphong@mpi-inf.mpg.de \quad shrestha.ghosh@uni-tuebingen.de}}
}

\begin{document}
\maketitle
\begin{abstract}
Large language models (LLMs) have majorly advanced NLP and AI, and next to their ability to perform a wide range of procedural tasks, a major success factor is their internalized factual knowledge. Since \citet{petroni-etal-2019-language}, analyzing this knowledge has gained attention. However, most approaches investigate one question at a time via modest-sized pre-defined samples, introducing an \textit{``availability bias''} \cite{kahnemann} that prevents the analysis of knowledge (or beliefs) of LLMs beyond the experimenter's predisposition.

To address this challenge, we propose a novel methodology to comprehensively materialize an LLM's factual knowledge through recursive querying and result consolidation. Our approach is a milestone for LLM research, for the first time providing \textit{constructive} insights into the scope and structure of LLM knowledge (or beliefs).

As a prototype, we build GPTKB, a knowledge base (KB) comprising 101 million relational triples for over 2.9 million entities from GPT-4o-mini. We use GPTKB to exemplarily analyze GPT-4o-mini's factual knowledge in terms of scale,
accuracy, bias, cutoff and consistency, at the same time. GPTKB is accessible at\\ \website.

\end{abstract}

\section{Introduction}

 

\begin{table}[t]
    \centering
    \small
    \begin{NiceTabular}{@{}p{0.44\columnwidth}|p{0.48\columnwidth}@{}}
    \toprule
    \textbf{Existing LLM analyses} & \textbf{Our proposal} \\
    \midrule
    \textcolor{red}{\ding{55}} \textbf{\textit{Prompt-and-discard:}}\newline
     - Transient single-focus explorations (e.g., factuality OR bias OR cutoff) & 
        \textcolor{green}{\checkmark} \textbf{\textit{Persistent resource:}} \newline
     - Materialized KB reusable for a wide array of questions
    \\[13pt]
        \textcolor{red}{\ding{55}} \textbf{\textit{Sample availability bias:}} \newline
     - Insights bound to experimenter’s predisposition
         & 
        \textcolor{green}{\checkmark} \textbf{\textit{Recursive materialization:}} \newline 
     - Discover unconceived LLM knowledge/beliefs \\[4pt]
    
     \textcolor{red}{\ding{55}} \textbf{\textit{Scratching the surface: }} \newline
     - Few 100s-100K samples \newline
     - Breadth and depth of LLM knowledge untouched
      &
     \textcolor{green}{\checkmark} \textbf{\textit{Massive-scale:}} \newline 
     - Over 100M records \newline - Recursive crawl to unprecedented breadth and depth
     \\
     \bottomrule
    \end{NiceTabular}
    \caption{Comparison of existing LLM knowledge analysis approaches and our proposal.}
    \label{tab:motivation}
\end{table}

LLMs have been one of the most exciting recent developments in NLP and AI, and next to their ability to perform a wide set of procedural tasks, a major success factor is the factual knowledge that they have internalized \cite{bubeck2023sparks}. Their potential to store factual knowledge, like \textit{(Nelson Mandela, award, Nobel Peace Prize)}, was first highlighted by \citet{petroni-etal-2019-language}, and this has generated an own (sub-)field of studying all aspects of factual knowledge in LLMs, from trying to locate it, to estimates of their storage potential, to techniques to effectively elicit it \cite{jiang-etal-2020-know,roberts-etal-2020-much,veseli2023evaluating,sun-etal-2024-head,wu2024towards}. A large set of benchmarks and studies transiently investigate the knowledge storage ability via example-based prompting, e.g., to determine, how many triples from common knowledge bases (KBs) or question answering (QA) datasets are known to LLMs. However, all these works are subject to an \textbf{availability bias} \cite{kahnemann}, i.e., they only surface LLM knowledge/beliefs\footnote{The terminology here is contentious, see Section \ref{epistemiology}.} on topics that were known to and pre-designed by the experimenter. They inherently cannot discover knowledge on topics outside of their benchmarks.

\begin{figure}[t]
 \centering
  \includegraphics[width=\columnwidth,]{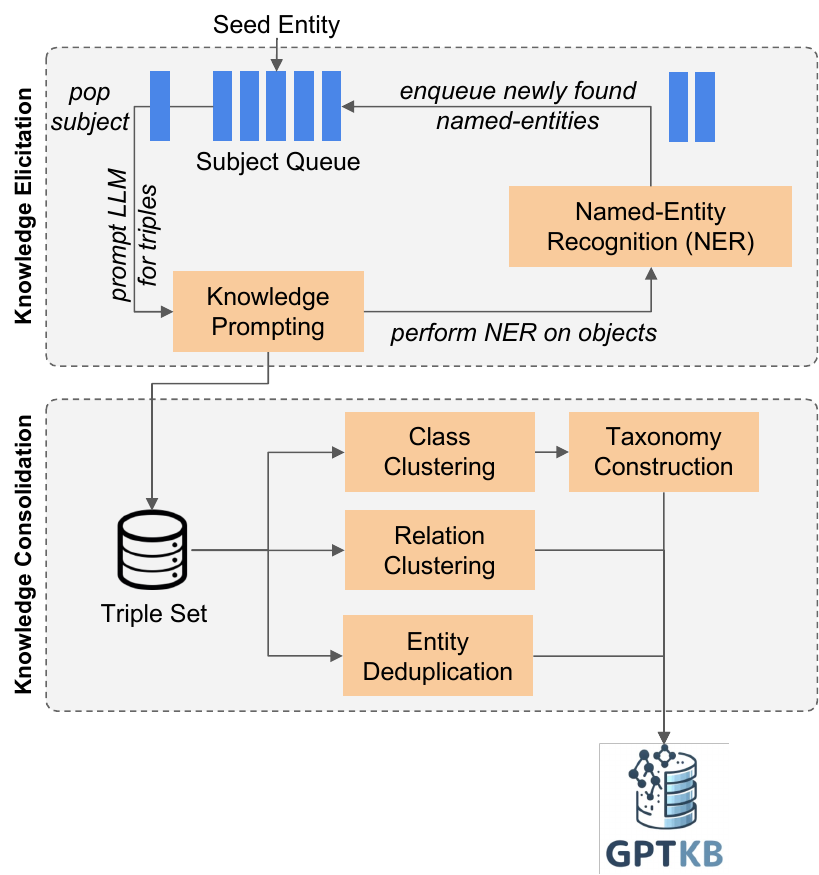}
 \caption{Overview of our approach for factual knowledge materialization from LLMs.
 }
 \label{fig:overview}
 \end{figure}

To enable comprehensive insights into LLM knowledge, we propose to extensively materialize the knowledge of LLMs into a KB (see Table~\ref{tab:motivation}). General-world KBs like Wikidata \cite{wikidata}, Yago \cite{yago} and DBpedia \cite{dbpedia} are important and long-standing backbones for AI applications, yet have seen little innovation in the last years, and have only been constructed manually, or by scraping semistructured web content \cite{machine-knowledge}.
In particular, we propose to use recursive graph exploration to obtain comprehensive (named-entity-centric) LLM knowledge, and to consolidate it via LLM-based entity disambiguation, class and relation canonicalization, and taxonomy induction (see Figure~\ref{fig:overview}). This proposal faces several challenges:
\begin{enumerate}
    \item \textit{Runtime and cost:} State-of-the-art LLMs potentially store millions of facts, and LLM inference is relatively slow. It is therefore unclear how to perform comprehensive knowledge elicitation under practical monetary and time constraints.
    \item \textit{Variance, hallucinations, and scoping:} Latent knowledge in LLMs covers a wide set of topics, varies by entity, and includes hallucinations. We need a method that elicits a high quantity, without encouraging hallucinations, and without falling into bottomless corners, e.g., open-ended phrases or translations. 
    \item \textit{Global inconsistency}: Consecutive prompts risks surfacing outputs that are not globally consistent, e.g., introducing duplicate relations, entities, or disconnected class hierarchies.
\end{enumerate}
Our approach builds on the following ideas: To overcome scaling issues, and obtain relevant knowledge, we utilize a commercial API that allows to massively batch requests, and utilize named entity recognition (NER) and carefully crafted prompts to restrict the explored space, along with prompts that encourage varied answer sizes. To obtain a coherent KB, we perform a set of canonicalization and disambiguation steps, entirely relying on the LLM itself. In summary, our salient \textbf{contributions} are:
\begin{enumerate}
    \item To the best of our knowledge, we are the first to propose  extensive materialization for analyzing factual LLM knowledge.
    \item We present a scalable methodology to elicit massive amounts of LLM knowledge, and to consolidate it.
    \item Using GPT-4o-mini, we construct \ourkb, the first large-scale materialization of LLM knowledge, comprising 101M assertions for over 2.9M entities. 
    \item We use \ourkb\ to exemplarily analyze GPT-4o-mini's factual knowledge in terms of scale, accuracy, bias, cutoff, and consistency, at the same time.
\end{enumerate}
We provide code and a concrete resource, \ourkb, both as a 3.8 GB download, and via an online browsing interface and SPARQL query interface at \website.

\section{Related work}

\paragraph{Extent of LLM knowledge}
Since the emergence of LLMs, the question of how much these models know has frequently been raised \cite{petroni-etal-2019-language,roberts-etal-2020-much,jiang-etal-2020-know,singhania2022lm,veseli2023evaluating,sun-etal-2024-head,wu2024towards}.
So far, the widely adopted approach is to sample a domain of interest, e.g., question-answer pairs from a common benchmark, or triples from Wikidata, transiently probe the LLM, and compute the fraction that the LLM can correctly answer/complete. 
This is prone to ``availability bias'' \cite{kahnemann} since we do not get a comprehensive view of the LLM knowledge outside the focus of existing benchmarks.
For instance, we found that GPT-4o-mini holds substantial knowledge about art periods or hobbies, which are not covered in existing KBs.
\citet{kassner-etal-2021-beliefbank} propose a small-scale persistent memory component for ensuring LLM answers remain consistent over multiple prompts.
\citet{mango} iteratively prompt GPT-3.5 for obtaining 167K sentences containing cultural commonsense for 11K subjects. \citet{cohen-etal-2023-crawling} propose to iteratively prompt GPT-3 for relations and relational assertions for triples.
In difference to \cite{mango}, we are after structured content (triples), at a much larger scale. In difference to \cite{cohen-etal-2023-crawling}, we unveil the practical challenges that such a graph exploration faces, tackle them, and perform LLM-based KB construction at scale.

\paragraph{Iterative information extraction}
Iterative information extraction is a long-standing idea, already prototyped by Google cofounder Sergey Brin via the DIPRE system \cite{brin1998extracting}. The Snowball system by \citet{agichtein2000snowball} presented a large-scale implementation of this idea, harvesting tuples from over 300,000 newspaper documents. The KnowItAll system \cite{etzioni2004web} followed a similar approach, and led to the ReVerb KB \cite{reverb}. Several recent works attempt IE \textit{with} LLMs from text \cite{hao2022bertnet,carta2023iterative,zhang2024automated}, but, except of the LM-KBC challenge series \cite{singhania2022lm} and a proposal by \cite{cohen-etal-2023-crawling}, to date, we are not aware of any attempts to extract knowledge \textit{from} LLMs.

\paragraph{Large-scale KB construction}
Dominating public large-scale KBs are Wikidata \cite{wikidata}, Yago \cite{yago} and DBpedia \cite{dbpedia}, all started more than 10 years ago. While Wikidata is constructed by volunteers, Yago and DBpedia represent the paradigm of (semi-)structured information harvesting and integration, extracting in particular from Wikipedia infoboxes and Wikidata \cite{machine-knowledge}. They all remain incomplete \cite{razniewski2024completeness}, warranting the search for novel paradigms. Commercial projects like the Google KG \cite{singhal2012knowledgegraph} or Amazon's KG \cite{dong2020autoknow} have usually followed these approaches. By comparison, text-based KB construction, e.g., in NELL \cite{nell} or ReVerb \cite{reverb}, has achieved less adoption. Our approach is more related to the latter approaches, as LLMs are distillations of large text corpora. 
Table~\ref{tab:KB-sizes} gives an overview of major KB projects. 

\begin{table}[t]
\small
\begin{center}
\begin{tabular}{@{}lrr@{}}
\toprule
\textbf{KB} & \textbf{\#entities} & \textbf{\#assertions} \\ \midrule
\textit{Wikimedia-related} \\
Wikidata & 113M & 1.62B \\
Wikidata5m & 5M & 20M \\
Yago 4.5 & 50M & 140M  \\
DBpedia & 3.8M & 75M \\ 
\midrule
\textit{Text-extracted} \\
NELL & ? & 12M \\
ReVerb & ? & 15M \\ 
\midrule
\textit{Generative} \\
\ourkb & 2.9M & 101M  \\ \bottomrule
\end{tabular}
\end{center}
\caption{Size comparison of major KBs. Sources in Appendix \ref{app:source-kb-sizes}.}
\label{tab:KB-sizes}
\end{table}


\section{LLM knowledge materialization}

An overview of our approach is shown in Figure~\ref{fig:overview}. In the first phase, \textbf{knowledge elicitation}, we iteratively elicit LLM triples for a given subject, and enqueue newly encountered named entities for further triple elicitation. In the second phase, \textbf{knowledge consolidation}, we consolidate the resulting triple set by canonicalizing entities, relations and classes, and by constructing a coherent taxonomy. 
Our paradigm refrains from imposing any standardized vocabularies, or using existing KBs as disambiguation references to effectively materialize the parameterized LLM knowledge.

\subsection{Knowledge elicitation}

\begin{figure}[t]
    \centering
    \includegraphics[width=\columnwidth]{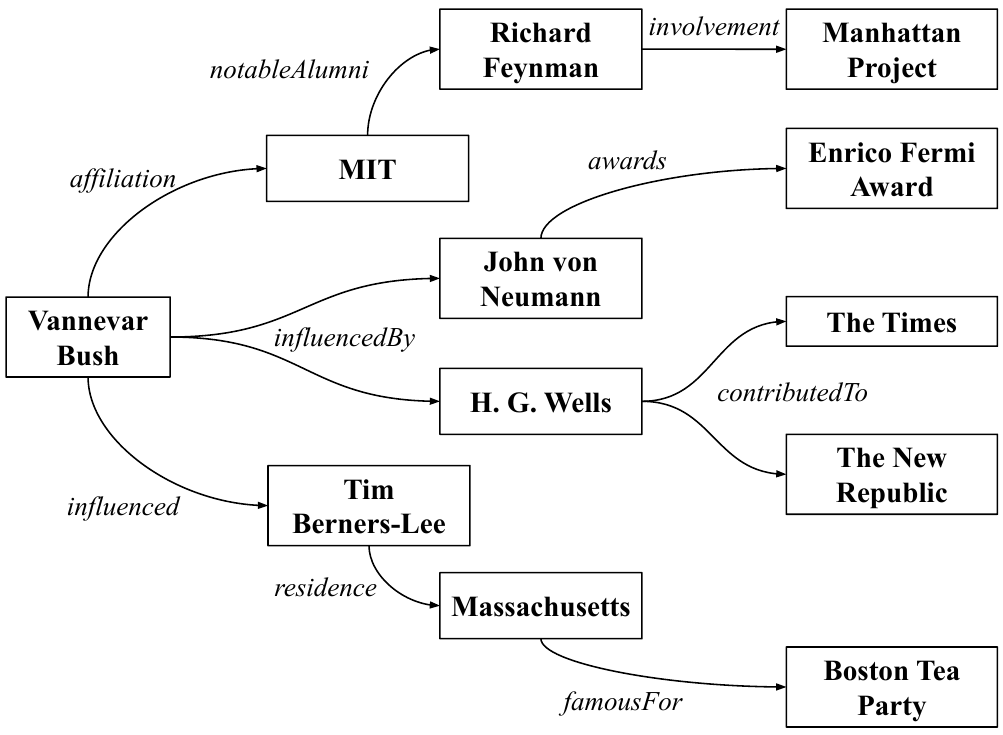}
    \caption{Graph exploration from the seed entity \textit{Vannevar Bush}.}
    \label{fig:graph_explore}
\end{figure}

\paragraph{Iterative graph expansion}
The process starts from one or a set of seed subjects, e.g., \textit{Vannevar Bush}, the visionary behind the concept of hyperlink-based knowledge organization \cite{bush1945we}.
We prompt the LLM to return knowledge about this subject in the form of triples.
From the objects in the triples obtained for him, we then identify further entities using NER (e.g., \textit{MIT} (affiliation) or \textit{Everett, MA} (birth place)). Thus, we iteratively expand the entity graph and elicit further knowledge on the newly encountered entities \cite{brin1998extracting}. We elaborate on the \textbf{\textit{knowledge prompting}} and \textbf{\textit{NER}} in the following paragraphs. Figure \ref{fig:graph_explore} illustrates how in 3 hops, we arrive at entities of diverse types, such as \textit{historical event} (Boston Tea Party, Manhattan Project), \textit{newspaper} (The Times), and \textit{magazine} (The New Republic).

\paragraph{Knowledge prompting}
A major challenge is to elicit as much knowledge as possible, but at the same time, not encourage hallucinations. We found that without guidelines on the expected number of triples, LLMs showed too little variance in the number of triples per subject, while we would expect them to return many more triples for \textit{Einstein} than for other entities. We solve this via indications that are defined in relation to the entity's popularity. In difference to the proposal \cite{cohen-etal-2023-crawling}, we also drop the separate relation elicitation, and relation-specific knowledge extraction, as these do not scale. To structure the knowledge, we also request that at least one \textit{instance\_of} triple should be returned. Output parsing is eased by using the structured outputs feature of OpenAI's API. In turn, this also reduces hallucinations, e.g., additional qualifiers, textual descriptions, or similar. The full prompt is in Figure~\ref{fig:full-prompt} in the Appendix.

\paragraph{Named-entity recognition (NER)}
Our early attempts at graph exploration were plagued by topical runaway into linguistic knowledge, translations, etc., because LLM-generated objects cover a wide range of imprecisely delineated data types. We experimented with various ways to filter non-named entities from the expansion, but found that existing NER frameworks struggled with the context-free entity labels available from our crawl process (most NER models are trained on sentences). In the end, we used the LLM itself to identify named entities, processing multiple candidates at once. The full prompt is presented in Figure~\ref{fig:prompt-ner} in the Appendix.


\subsection{Knowledge consolidation}


The knowledge elicitation phase 
returns
a huge degree of redundancy and variance.
Feeding a large amount of existing entities or taxonomy into the knowledge elicitation prompt could mitigate this problem.
However, that would also increase costs significantly, and even be infeasible once we have generated millions of entities and triples.
Instead, we introduce several steps to consolidate LLMs output post-hoc.

\paragraph{Relation clustering}
The elicitation phase
generates
788K
distinct relation names, with many obvious duplicates, e.g., \textit{instanceOf}, \textit{isA}, or \textit{InstanceOf}. 
We apply a greedy clustering algorithm to this set (see Algorithm~\ref{alg:relation-clustering}).
%
%
This process merges relation $r$ into a more frequent relation $s$, selected as the one with the highest textual embedding similarity to $r$ among all relations more frequent than $r$, if the similarity is greater than an adaptive threshold (defined in line 6 in Algorithm~\ref{alg:relation-clustering}).
This
threshold varies with the frequency of the relation, leading to a more aggressive removal of relations 
with low frequency. 

\vspace{-0.1cm}
\begin{algorithm}[ht]
\small
\caption{Relation clustering}\label{alg:relation-clustering}
\begin{algorithmic}[1]
\Require A set of relations $\mathcal{R}$, hyperparameter $\alpha$
\Ensure A map from relation to cluster-id $\mathcal{C}$
\State $\mathcal{R} \gets \text{sort}(\mathcal{R}, \text{key} = \text{frequency}, \text{reverse} = \text{True})$ 
\State $\mathcal{C} \gets \{\}$
\State $\textit{next\_cluster\_id} \gets 0$
\For{$r \in \mathcal{R}$}
    
    \State $s' = \text{argmax}([\text{similarity}(r, s) \textbf{ for } s \in \mathcal{R} \textbf{ upto } r])$
    \State $\textit{threshold} \gets \text{clip}\{\alpha \times \frac{\log(\text{frequency}(r))}{\log(\text{frequency}(\text{first}(\mathcal{R})))}, \mathcal{H}, \mathcal{L}\}$

    \If{$\text{similarity}(r, s') > \textit{threshold}$}
        \State $\mathcal{C}[r] \gets \mathcal{C}[s']$
    \Else
        \State $\mathcal{C}[r] \gets \textit{next\_cluster\_id}$ 
        \State $\textit{next\_cluster\_id} \gets \textit{next\_cluster\_id} + 1$
    \EndIf
    
\EndFor
\State \Return $\mathcal{C}$
\end{algorithmic}
\end{algorithm}
\vspace{-0.1cm}

\paragraph{Class clustering}
Similarly to relations, there are 103K distinct class names
(i.e., objects for \textit{instance\_of} relations) 
generated by the knowledge elicitation phase, 
with both obvious duplicates, and many overly specific cases. 
Algorithm~\ref{alg:relation-clustering} is also applied to clean this set.

In our experiment, both relation and class clustering uses SentenceTransformers \cite{reimers-2019-sentence-bert} for embedding cosine similarity computation, with $\alpha = 1.4$ and global highest threshold $\mathcal{H} = 0.95$ chosen via manual inspection of small held-out sets. For tail predicates, to ensure clustering precision, we set lowest threshold $\mathcal{L} = 0.75$.

\paragraph{Taxonomy construction}
The classes 
in the KB
so far 
do not form 
a 
coherent, or even 
connected taxonomy, 
as the knowledge elicitation process only expands named entities. 
But we also cannot simply prompt for a complete taxonomy to put on top (e.g., following the approach by \citet{funk2023towards}), because we need to include the existing classes. 
We propose Algorithm~\ref{alg:taxonomy-construction}, which is based on the LLM itself, to build a complete taxonomy that incorporates given classes.

\begin{algorithm}[ht]
\small
\caption{Taxonomy construction}\label{alg:taxonomy-construction}
\begin{algorithmic}[1]
\Require Knowledge base (KB)
\Ensure Taxonomy
\Function{construct\_taxonomy}{KB}
    \State \textit{taxonomy} $\gets$ LLM\_get\_high\_level\_taxonomy()
    \For{\textit{class} $\in$ KB.classes} 
        \State \textit{class}.generality\_score $\gets$ LLM\_get\_score(\textit{class})
    \EndFor
    \State \textit{classes} $\gets$ sort(KB.classes, key=generality\_score)
    \For{\textit{class} $\in$ classes} 
        \State insert\_class\_recursive(\textit{taxonomy}.root, \textit{class})
    \EndFor
    \State \Return \textit{taxonomy}
\EndFunction
\Statex
\Function{insert\_class\_recursive}{\textit{node}, \textit{class}}
    \If{has\_children(\textit{node})}
        \State \textit{next} $\gets$ LLM\_superclass(\textit{class}, \textit{node}.children)
        \If{\textit{next} \textbf{is not} $\textsc{null}$}
            \State insert\_class\_recursive(\textit{next}, \textit{class})
        \Else
            \State LLM\_update\_taxonomy(\textit{node}, \textit{class})
        \EndIf
    \Else
        \State LLM\_update\_taxonomy(\textit{node}, \textit{class})
    \EndIf
\EndFunction
\end{algorithmic}
\end{algorithm}

The construction starts by generating a high-level taxonomy via the LLM.
Then, for each of the existing classes in the KB (sorted by generality scores given by the LLM), we find the lowest node of which the given class is identified as a subclass, via depth-first search (see function \textsc{insert\_class\_recursive} in Algorithm~\ref{alg:taxonomy-construction}).
Then, we ask the LLM to update the sub-taxonomy starting from the found node with the given class.
In this step, the LLM may generate new intermediate classes on the path from the found node to the given class.

The prompts used for seed taxonomy generation, generality scoring, superclass identification, and taxonomy update are provided in Figure~\ref{fig:taxonomy} in Appendix \ref{app:prompts}.
Further LLM-based refinement in the style of \cite{peng2024refining} could be considered.

\paragraph{Entity deduplication}

Naive graph exploration frequently yields duplicates, 
such as
\textit{John F. Kennedy} and \textit{John Fitzgerald Kennedy}. To remove these duplicates without requiring extraordinary runtime, we follow the standard blocking-based deduplication approach by \citet{kopcke2010frameworks}. 
This 
approach
is 
based on choosing a blocking key,
which is an
attribute by whose values we obtain meaningful partitions on entities. 

We focus on entities that are instances of the most interesting class, \textit{humans}. We choose to block entities by birth dates.
Within each block, we consider an entity pair as duplicates if 
(i) their labels 
have close meanings (i.e., cosine similarity between SentenceTransformers embeddings are greater than 0.85),
and 
(ii) 30\% of their triples are exactly the same. 
More advanced methods could utilize LLMs themselves for deciding on entity equivalence \cite{ding2024entgpt}, but it would incur more cost and longer runtime.

\section{Implementation}
\label{sec:implementation}

\paragraph{LLM choice and parallelization}
We chose \textit{GPT-4o-mini} \cite{openai2024gpt4technicalreport} for our experiments, because of (i) its ability to process requests in batches, (ii) its structured output feature, (iii) its good tradeoff between performance and cost. The batch feature is particularly important, enabling us to send, after startup, up to 100 batches of 10,000 entities in parallel. The model's size is not publicly released, but has been estimated to be around 8B \cite{gptparameters1,gptparameters2}.

\paragraph{Seed entity, runtime and cost} We start the process from a single entity, \textit{Vannevar Bush}, in honor of his vision of interlinked knowledge \cite{bush1945we}. 
Note that this choice is arbitrary and overall inconsequential.
As general-world knowledge graphs are densely connected \cite{hogan2021knowledge}, any seed entity connected to popular entities serves as well.
For instance, we reach \textit{MIT} in one hop, \textit{Alan Turing} in 2 hops, \textit{Kyoto} in 3 hops.

We require a total of 2,200 batches to prompt the LLM up to BFS-depth 10, which corresponded to 5.8M prompted entities, of which for 2.9M a non-empty answer was returned. The whole process took 27 hours, and including trial runs, constructing \ourkb\ cost us a total of \$3,500 for 
API calls.

\paragraph{\ourkb{} statistics}
Our KB contains a total of 101M triples for 2.9M entities, organized into 567K relations and 4,715 classes. This gives an average of 35 triples per subject, with two distinct clusters, in particular, 860K entities with 10 triples and 66K with 50 triples, with most others near these values. Of all triples, more than 37M have an entity as object, and more than 64M have a literal as an object. The average length of entity labels is 21.8 characters. Example output for \textit{Vannevar Bush} is shown in 
Figure~\ref{fig:graph_explore}.
Statistics are shown in Table~\ref{tab:statistics}.

\begin{table}[t]
\centering
\resizebox{\columnwidth}{!}{%
\small
\begin{tabular}{@{}ll@{}}
\toprule
\textbf{Entities} & 2.9M \\
\textbf{Triples} & 101M \\
\textbf{Relations} & 567K (788K before canonicalization) \\
\textbf{Classes} & 4,715 (103K before canonicalization) \\
\textbf{Triple objects} & 37M entities, 64M literals \\
\textbf{Avg. triples/entity} & 35 \\
\textbf{Avg. label length} & 21.8 characters \\
\textbf{Avg. outlinks} & 4 \\
\textbf{Subject-precision*} & 74\% Verifiable, 9\% Plausible,\\
 & 17\% Unverifiable \\
\textbf{Subjects in Wikidata*} & 37\% in WD, 63\% not in WD \\
\textbf{Triple-precision*} & 31\% True, 61\% Plausible, \\
                         & 1\% Implausible, 7\% False \\ 
\bottomrule
\multicolumn{2}{l}{\small\textbf{*} \textit{as a weighted average across layers.}}
\end{tabular}
}
\caption{Statistics of \ourkb. }
\label{tab:statistics}
\end{table}

\paragraph{Dataset provision and license} We provide our KB as a download (3.8 GB in TTL format), via a web browsing interface, and via a SPARQL query endpoint at \website \ under the permissive CC BY-NC 4.0 license, though we caution against unreflected downstream usage (see Section~\ref{sec:limitations}). This license is compatible with the permissive terms of use of OpenAI \cite{openai2022sharingpublicationpolicy}.

\section{Analysis}
\label{sec:discussion}

\subsection{Precision}
\label{sec:precision}

Evaluating the precision of large-scale KBs is not straightforward, because they contain a significant amount of long-tail knowledge, for which finding evidence or counter-evidence is difficult. In line with
\citet{yago}, we evaluate precision within the context of web-retrievable information. Figure \ref{fig:accuracy} summarizes our findings.

For verifying \textbf{entities}, we retrieved 5 search results from a web search API for the entity label, which are then used to decide whether the entity's existence could be verified, appears plausible, or could not be verified. Deciding on these labels based on textual context represents a task of textual entailment, a.k.a. natural language inference (NLI), a task generally considered solved for LLMs.\footnote{On LLM performance on textual entailment, the popular SNLI \cite{bowman-etal-2015-large} is comparable, which was considered solved in 2020 by BERT architectures \cite{bert-snli} at 92\% accuracy, which is the ceiling, given dataset noise. This is also why the literature mentions no numbers for newer models like Llama on this task anymore.} We therefore use another LLM, Llama-3.1-70B-Instruct~\cite{llama31}, for this task. The LLM hereby only judges textual entailment, it does not independently judge truth from its parameters.
Based on 1,000 samples, average entity verifiability is 74\%, and as shown in Figure~\ref{fig:accuracy}, we observe a continuous drop of verifiability over the layers, stabilizing at layer 10 at about 70\%. The plot also includes Wikidata-existence, which paints a similar trend at lower levels. We generally observe a wide variance across classes, finding, for instance, that persons consistently have a higher verifiability than fictional characters, where many codes were made up (e.g., \textit{Officer K.I.T.T. XV} and \textit{Officer K.I.T.T. XVI}) or varying incorrect type numbers are added to real names continuously by the LLM (e.g., \textit{Agent 71} and \textit{Agent 72}). 

\begin{figure}[t]
    \centering
    \resizebox{\columnwidth}{!}{%
    \begin{tikzpicture}
    \begin{axis}[
        xlabel={BFS-Layer},
        ylabel={Percentage (\%)},
        ymin=0, ymax=100,
        xmin=1, xmax=10,
        xtick={1,2,3,4,5,6,7,8,9,10},
        grid=both,
        width=10cm, height=6cm,
        legend style={
                at={(0.5,-0.3)}, 
                anchor=north,    
                cells={align=left}, 
                draw=none
            },
        legend columns=2,
    ]
    \addplot[
        very thick,
        color=gg-red,
        mark=square*,
        mark options={scale=0.8}
    ] coordinates {
    (1,100) (2,100) (3,93) (4,87) (5,80) (6,78) (7,66) (8,69) (9,71) (10,69)
};
    \addlegendentry{Entities web-verifiable}

    \addplot[
        very thick,
        color=gg-orange,
        mark=*,
        mark options={scale=0.8}
    ] coordinates {
    (1,100) (2,100) (3,76) (4,65) (5,52) (6,42) (7,29) (8,20) (9,26) (10,23)
};
    \addlegendentry{Entities in Wikidata}
    
    \addplot[
        very thick,
        color=gg-blue,
        mark=triangle*,
        mark options={scale=0.8}
    ] coordinates {
        (1,62) (2,58) (3,60) (4,61) (5,34) (6,31) (7,30) (8,25) (9,36) (10,27)    };
    \addlegendentry{Triples web-verifiable}
    
    \addplot[
        very thick,
        color=gg-teal,
        mark=diamond*,
        mark options={scale=0.8}
    ] coordinates {
        (1,20) (2,33) (3,32) (4,36) (5,58) (6,63) (7,61) (8,67) (9,56) (10,63)
    };
    \addlegendentry{Triples web-plausible}
    \end{axis}
    \end{tikzpicture}
    }
    \caption{Accuracy of \ourkb\ across layers.}
    \label{fig:accuracy}
\end{figure}

For a given \textbf{triple}, we retrieve the top-5 web-page snippets for the search term \textit{<subject> <object>} from a search engine API, then again perform LLM-based textual entailment to decide whether the given triple is entailed (entailed/plausible/implausible/false), given the snippets. We specifically adapt the prompt used by \citet{adam2024traceable}, extending it with the ``false'' level. Based on 1,000 samples, we find that on average, 31\% of triples are entailed, 61\% are plausible, 1\% is implausible, and 7\% are false.

In terms of the accuracy of the \textbf{taxonomy}, we follow the per-edge evaluation scheme proposed by \citet{bordea2016semeval}, finding that 64\% of all subclass-superclass edges are considered correct, using the Llama model as judge. We also evaluate whether a superclass is most appropriate, by offering all siblings as alternatives. Here, for 70\% of all subclass-superclass edges, the superclass is considered the best alternative. Note that the structure of our taxonomy still leaves room for improvement in terms of long-range dependencies and distributions, that are not easy to quantify or address locally.


\begin{figure}[t]
\centering
\scriptsize
\begin{tikzpicture}
    \pie[sum=auto, after number=K, radius=2]{
    281/Person, 
    66/film, 
    56/Company, 
    46/video game, 
    43/album, 
    37/museum, 
    33/fictional character,
    32/character, 
    31/City,
    30/song
    }
\end{tikzpicture}
\caption{The 10 most frequent classes in \ourkb, which make up 21.5\% of all entities.}
\label{fig:frequent_classes}
\end{figure}

\subsection{\ourkb{} content and comparison}
\label{sec:content-and-comparison}

\begin{figure}[t]
 \centering
  \includegraphics[width=\columnwidth,trim=0.8cm 0 0.7cm 0]{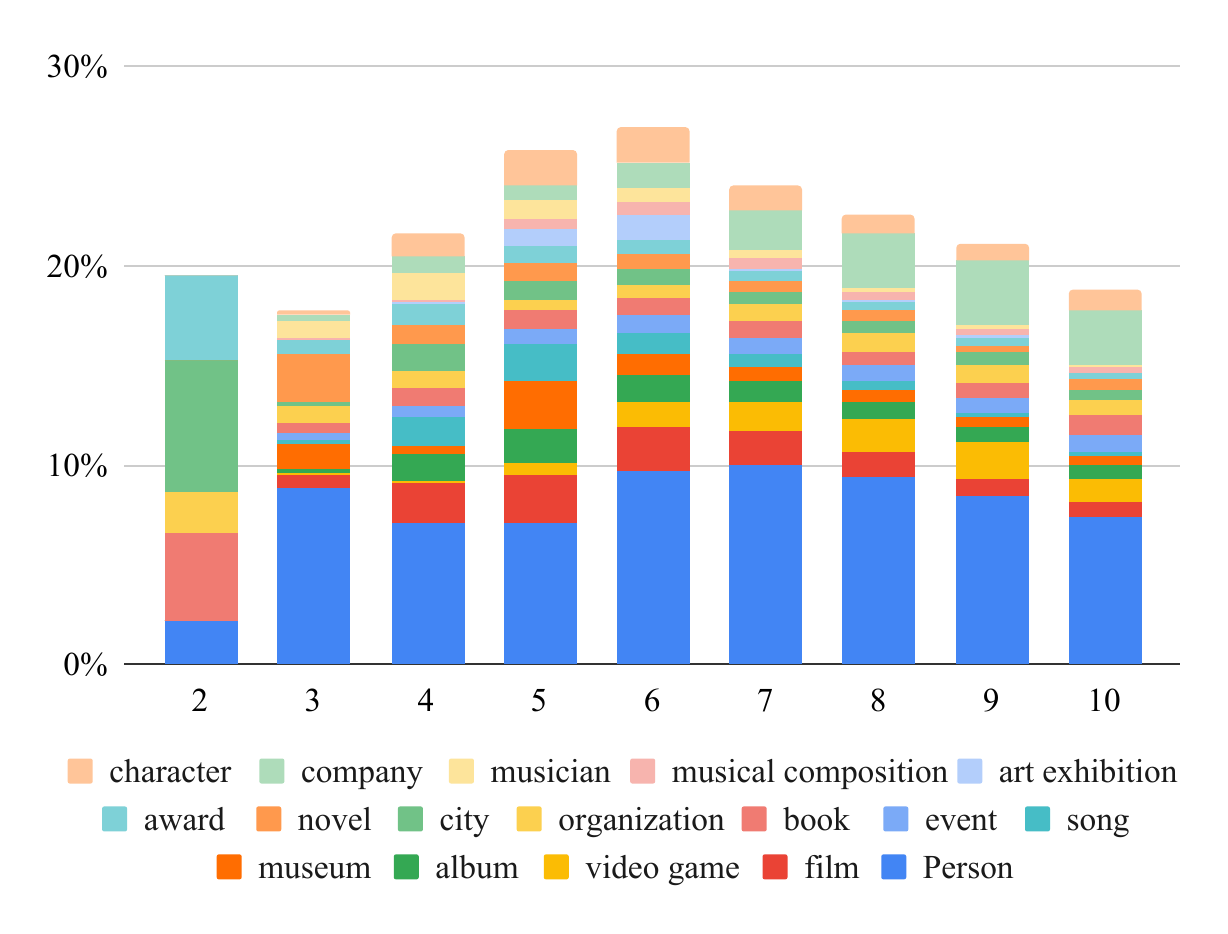}
 \caption{Most frequent classes by BFS-level. The root level (\#1) is excluded from the chart.}
 \label{fig:classes-by-bfs-level}
 \end{figure}

\paragraph{Classes} The most frequent classes generated for entities are shown in Figure~\ref{fig:frequent_classes}, where we find that the class \textit{Person} dominates the dataset with 281K instances. This is followed by the classes \textit{film} (66K) and \textit{Company} (56K), which are are an order of magnitude smaller. Classes denoting creative works, such as, \textit{video game} (46K), \textit{album} (43K) and \textit{song} (30K), also make it to the top ten frequent classes.
Figure~\ref{fig:classes-by-bfs-level} shows the most frequent classes by BFS-level. As one can see, diversity increases towards lower BFS-levels, and some classes, like \textit{awards}, only emerge later.


\paragraph{Properties} The most frequent properties in the whole KB are \textit{instance\_of} (3.1M) and \textit{features} (2M) with similarly generic properties following at the entity level (see Figure~\ref{fig:frequent_properties_all}). At the class level, we find more specific properties, for instance, the most frequent properties in the class \textit{Person} are \textit{instance\_of} (299K), \textit{occupation} (191K), \textit{known\_for} (176K), \textit{nationality} (166K), and in the class \textit{film}, the properties \textit{character} (399K), \textit{starring} (134K) and \textit{genre} (99K) are the most frequent, shown in Figures~\ref{fig:frequent_properties_person} and \ref{fig:frequent_properties_film}, respectively.

\begin{figure*}[t]
\centering
\tiny
    \begin{subfigure}{0.32\textwidth}
        \centering
        \begin{tikzpicture}
            \pie[sum=auto, after number=M, radius=1.5]{
            3.0/instance\_of,
            1.9/features,
            1.3/associated\_with,
            1.0/is\_part\_of,
            0.9/is\_known\_for,
            0.8/location,
            0.8/character,
            0.7/offers,
            0.7/influenced\_by,
            0.7/provides
            }
        \end{tikzpicture}
        \caption{\ourkb{}.}
        \label{fig:frequent_properties_all}
    \end{subfigure}
    \hfill
    \begin{subfigure}{0.32\textwidth}
        \centering
        \begin{tikzpicture}
            \pie[sum=auto, after number=K, radius=1.5]{
            299/instance\_of,
            191/occupation,
            176/known\_for,
            166/nationality,
            161/birth\_place,
            142/notable\_work,
            139/exhibition,
            95/birth\_date,
            75/has\_occupation,
            70/spouse
            }
        \end{tikzpicture}
        \caption{Class \textit{Person}.}
        \label{fig:frequent_properties_person}
    \end{subfigure}
    \hfill
    \begin{subfigure}{0.32\textwidth}
        \centering
        \begin{tikzpicture}
            \pie[sum=auto, after number=K, radius=1.5]{
            399/character,
            134/starring,
            99/genre,
            79/influenced\_by,
            79/instance\_of,
            66/language,
            64/cast,
            58/features,
            58/runtime,
            57/theme
            }
        \end{tikzpicture}
        \caption{Class \textit{film}.}
        \label{fig:frequent_properties_film}
    \end{subfigure}
\caption{Distribution of 10 most frequent properties in \ourkb, and within classes \textit{Person} and \textit{film}.}
\label{fig:frequent_properties}
\end{figure*}

\paragraph{Content bias} 
We also analyze the geographic bias of our KB, using the listed nationalities as proxy. 
We 
observe a clear bias towards English-language nationalities: 
the top-2 are
American (119K) and
British (35K),
while the next 3 are French (18K), German (14K), and Japanese (11K), and Chinese and Indian only at 3K and 7K.
This bias in \ourkb{} 
is stronger than, e.g., in Wikidata \cite{shaik2021analyzing}, likely reflecting the (undisclosed) English-centric training corpus of GPT-4o-mini. 

An interesting point is also gender bias: The \textit{gender} property of \ourkb{} contains 15K female versus 8K male values. For first names, male ones are still more than female (47\% vs. 37\% based on the \textit{gender-guesser 0.4.0} Python library), but this is still a much lower bias than in other KBs, likely reflecting efforts of gender debiasing the LLM.


\paragraph{Wikidata overlap and novelty}
We compare with Wikidata \cite{wikidata} on several aspects: \textbf{First}, we compute the fraction of subjects that exist in Wikidata. From a random sample of 2,000 subjects from \ourkb, 24\% have an entity with exactly matching label in Wikidata. A further 6.5\% have a non-empty search result, i.e., an entity of paraphrased or similar label. The remaining 69.5\% appear novel.  
\textbf{Second}, we exemplarily look at the 41 triples for \textit{Vannevar Bush}, of which we find that more than 10 are not contained in Wikidata, e.g., his affiliation with the US government, his children count (incorrect by 1), or him inventing the concept of hypertext.\footnote{An exact comparison is not straightforward, because relation names do not perfectly align, and objects in \ourkb\ are substantially more wordy.} \textbf{Third}, we identify several properties not modelled at all in Wikidata, for instance, \textit{historical\_significance} (342K triples), \textit{art\_style} (84K triples), or \textit{hobbies} (24K triples). 
This indicates that \ourkb\ potentially contains a significant amount of novel knowledge. A more comprehensive KB comparison in the style of \citet{farber2018linked} is conceivable.


%

\section{Discussion}

\subsection{Lessons for LLM epistemology}
\label{epistemiology}

\paragraph{Terminological observations}
The notion and definition of LLM knowledge itself is controversial \cite{fierro2024defining}. We adopt here the term \textit{knowledge base} because it is common in the field, but would consider the term \textit{belief base} equally appropriate. In terms of the definitions of knowledge that are introduced by \citet{fierro2024defining}, we find that our output falls into the minimalistic sui-generis (g-knowledge) category, as our KB contains false as well as inconsistent statements. Beyond the reported errors, e.g., in many cases, spouse relations are not symmetrically represented (see ``enabled analyses'' below). 

\paragraph{Storage capacity}
Scaling laws and storage capacity are intensively investigated \cite{allen2024physics}, but usually on synthetic data. Our experiments provide a lower bound on real data: As discussed in Section \ref{sec:implementation}, GPT-4o-mini likely has in the order of 8B parameters. Given the 101M triples that we obtained, that makes \textit{79 parameters per triple}.\footnote{From \cite{allen2024physics}, a theoretical lower bound of 4 parameters/triple can be deduced, based on Remark 4.4, and their synthetic dataset containing 6 triples per entity. However, the generalizability of results on that synthetic dataset is unclear.} We emphasize that this number captures only encyclopedic knowledge, yet that LLMs also possess other knowledge, e.g., linguistic knowledge, procedural knowledge. 

\paragraph{Knowledge consistency}
To assess the consistency of LLM knowledge, we repeated the elicitation 100 times on the entity \textbf{Vannevar Bush}. From the perspective of knowledge volume consistency, the results primarily fall into two clusters: one comprising 52 runs with a mean of 21 triples (std = 1), and another with 32 runs averaging 38 triples (std = 6). The remaining 10 runs failed to parse. In terms of knowledge content consistency, within the first cluster, we observe 1,116 total triples and 79 unique ones, i.e., each triple is shared on average by 14 runs. The average overlap between the sets measured using an exact match set intersection is 0.67. Semantically, these numbers represent a lower bound only, as responses sometimes contain paraphrases. 

\paragraph{Complementarity} We observe a high complementarity to existing resources, with 63\% 
of generated entities not being present in Wikidata. Particularly, from Figure \ref{fig:accuracy}, \ourkb's entities remain web-verifiable even in deeper layers, about 70\% of the entities are web-verifiable in layer 10, even though the percentage of entities found in Wikidata drops to a little over 20\%.
Some particularly noteworthy complementary slices of LLM knowledge concern digital media artifacts, art periods, and people's hobbies.

\paragraph{Accuracy}
While on the entity level, most entities appear to truly exist (74\% verifiable, 9\% plausible), on the triple level, results are lower (31\% verifiable, 61\% plausible). This indicates that the LLM generally has a grasp of entities, but has more difficulties in correctly modeling their relationships. BFS-depth and quality are negatively correlated, but there is no complete quality collapse within the first 10 levels. 

\paragraph{Enabled analyses}
Our materialized resource enables a range of analyses about factual LLM knowledge, which for space reasons we only skim here:  \textbf{1) Accuracy}: See Sec.~\ref{sec:precision} and paragraph above. \textbf{2) LLM bias}: See Sec.~\ref{sec:content-and-comparison}. 
\textbf{3) Timeliness}: Exemplarily, we collect the most frequently mentioned years, and observe that there is sharp drop after 2023, which matches the knowledge cutoff of the LLM (see plot in Appendix~\ref{app:years}).
\textbf{4) Subject-wise consistency:} Our KB still contains many duplicates, e.g., \textit{The Elbe River, River Elbe, Elbe River, river Elbe, Elbe}. Studying their triples gives insights into consistency, e.g., we observe a significant semantic overlap, but also frequent different wordings (e.g., \textit{wildlife - various fish species / fish species}), and minor factual deviations (e.g., length 3x 1094km, 2x 1091km).
\textbf{5) Structural consistency:} An interesting observation concerns the difficulty that LLMs have with inverse relations \cite{reversalcurse}. We can observe this too, for instance, out of 318K \textit{spouse} triples, only 8K are symmetric, and out of 61K \textit{parent\_company} triples, only 6K are mirrored in \textit{subsidiary} triples.


\subsection{Lessons for knowledge materialization}

The \ourkb\ prototype reveals several important lessons for knowledge materialization. In particular, we find it notable that building such a large KB was possible so quickly, with a relatively small model.

\paragraph{Precision-recall trade-off} The biggest challenge in our view is precision, both in terms of hallucinated entities, and triples. We do not expect that larger models alone will solve this problem, because the long tail, where hallucinations occur, would likely just be pushed farther, but remain difficult to delineate. Tuning the precision-recall tradeoff, for example, via more conservative prompts, or via thresholding based on elicited confidence values \cite{xiongcan}, might be a way forward.

\paragraph{Emerging entities} Some applications are especially interested in newly emerging entities, and these are a long-standing challenge for traditional KBC \cite{knowledge-awakens}. Web-scraped KBs like Yago and DBpedia could in principle re-run their scrapers periodically, while text-extracted KBs like Nell and ReVerb would require re-runs on new web crawls. Our approach could proceed analogous, re-running the materialization on a newer version of the utilized LLM. If one were to know which entities are affected by updates, one could also perform retrieval-augmented generation, however, knowing where updates occured essentially requires an oracle, and none of the existing KBs successfully employed selective updating.

\paragraph{Consolidation challenges} Our exploration surfaced a potpourri of other challenges, some of them known to KB construction research since years \cite{machine-knowledge}, others requiring novel adaptations in the light of LLMs. These concern (1) NER for short labels without context, (2) entity deduplication, (3) entity canonicalization, (4) literal typing and canonicalization, (5) relation clustering, (6) relation organization in terms of subrelations, (7) class clustering, (8) taxonomy construction, and (9) triple verification. While we explored simple techniques for 1, 2, 5, 7, 8, each has room upward, and other tasks were not treated so far.

\paragraph{Cost-effectiveness} The \ourkb\ approach deviates from traditional Wikimedia- and data-integration focused KB construction, and although its precision still needs improvement, it stands out with its potential for cost efficiency. In a back-of-the-envelope calculation,
\citet{paulheim2018much} estimated the cost per triple for existing manual KB construction projects at \$2-6, and for existing automated KB construction projects at \$0.01-0.15. In contrast, in our prototypical execution, the API cost is just \$0.0001 per correct triple (\$3.5K/33M), i.e., $>$100x less than with previous methods.

\subsection{Other LLMs}
We performed a parallel exploration using Llama-3.1-70B-Instruct on local HPC hardware, and while accuracy is higher (69\% of triples verified true in a test run), this did not scale. We also envisioned a run using the strongest publicly available LLM, GPT-4o (80\% triples verified true), however, by its 15x higher API cost, and its estimated 25x larger parameter set, assuming knowledge is roughly proportional to parameters, at about \$825K, this is beyond our budget. 

In Table~\ref{tab:llm-comparison}, we compare GPT-4o-mini with GPT-4o and Llama 3.1 70B in terms of accuracy. All LLMs are evaluated on a similar-sized set of entities (20K). One can observe significant accuracy differences, consistent with the order of their (estimated) parameter size, and general benchmark results. 

\begin{table}[H]
\small
\centering
\begin{tabular}{lrrr}
    \toprule
     & \textbf{GPT} & \textbf{Llama} & \textbf{GPT} \\
     & \textbf{4o-mini*} & \textbf{3.1 70B} & \textbf{4o} \\
     \midrule
    Web-verified triples & 0.38 & 0.69 & 0.78 \\
    Entities on Wikidata & 0.78 & 0.83 & 0.88 \\
    Web-verified entities & 0.80 & 0.95 & 0.98 \\ 
    \bottomrule
\end{tabular}
\caption{Entity and triple verifiability comparison between different LLMs on a similar-sized set of entities. GPT-4o-mini* corresponds to the first 5 layers of what is reported in Figure~\ref{fig:accuracy}.}
\label{tab:llm-comparison}
\end{table}

\section{Conclusion}



We propose a novel methodology to recursively materialize LLM knowledge. For the first time in NLP and LLM research, our work provides comprehensive insights into what LLMs know (or believe). 
Our resource
is accessible at \website.

\section{Limitations}
\label{sec:limitations}


\paragraph{Prompt dependency} As already observed by \citet{petroni-etal-2019-language}, prompt formulations influence LLM responses, and hence  the resulting KB. \ourkb\ presents one way of materializing LLM knowledge into structured format, but better prompts may exist. In line with \citet{petroni-etal-2019-language}, methods like ours therefore always inform about lower bounds (on LLM factual knowledge, in our case).



\paragraph{Reproducibility} Our materialized KB is based on a closed-source LLM that does not come with guarantees regarding persistent servicing, and in the past, similar services have been discontinued, hence, long-term reproducibility is not guaranteed. At the same time, our resource therefore has especial archival value. An execution based on an open model is planned.

\section*{Acknowledgment}
We thank Markus Krötzsch and Gerhard Weikum for comments on this work, and Moritz Müller for help with the website.

\bibliography{custom}


\appendix

\section{Further KB Content}

\begin{figure}[ht]
\includegraphics[width=\columnwidth]{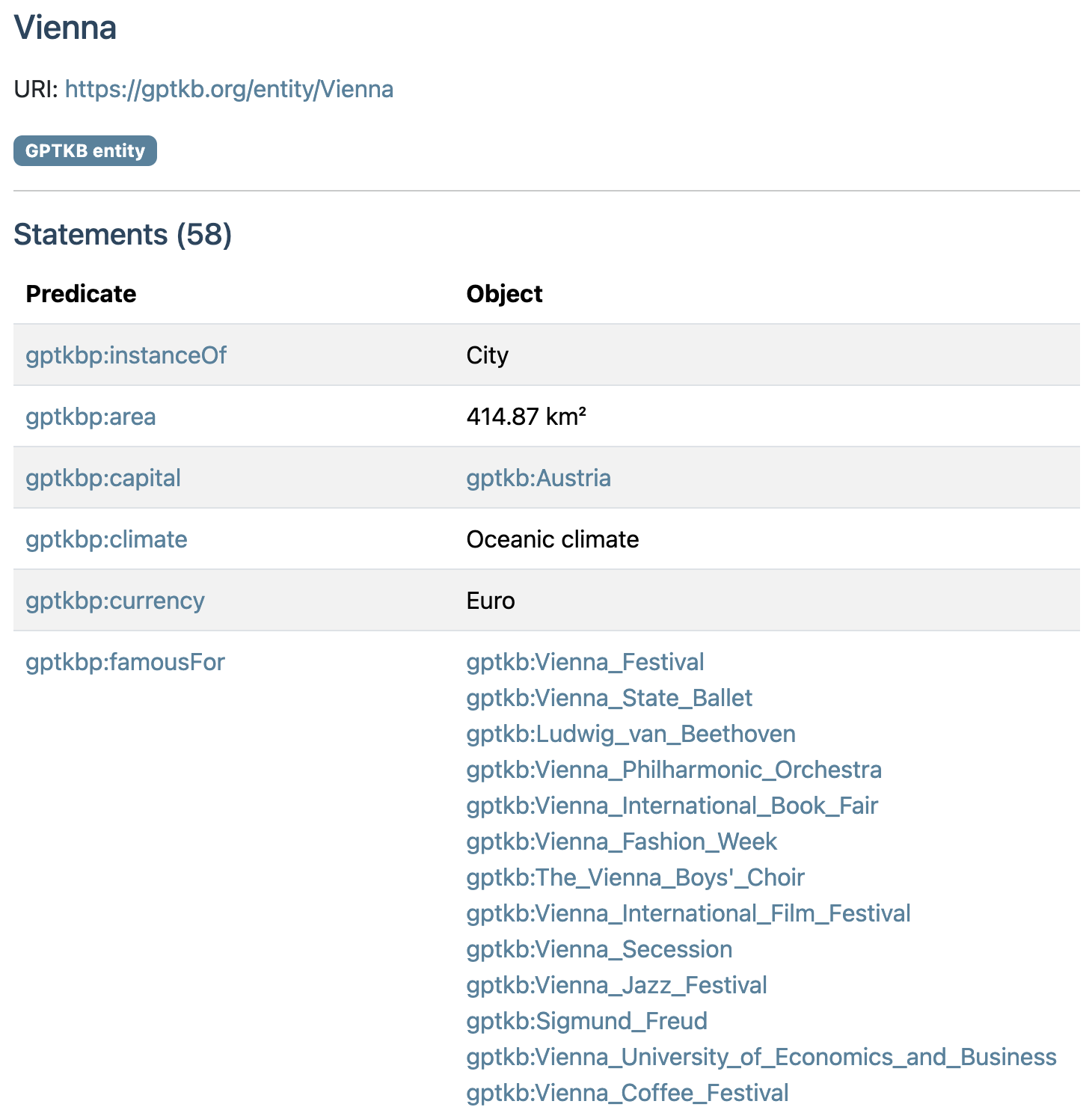}
\caption{Excerpt of \ourkb\ triples for subject \textit{Vienna}.}
\label{tab:vienna}
\end{figure}

\begin{figure}[ht]
\includegraphics[width=\columnwidth]{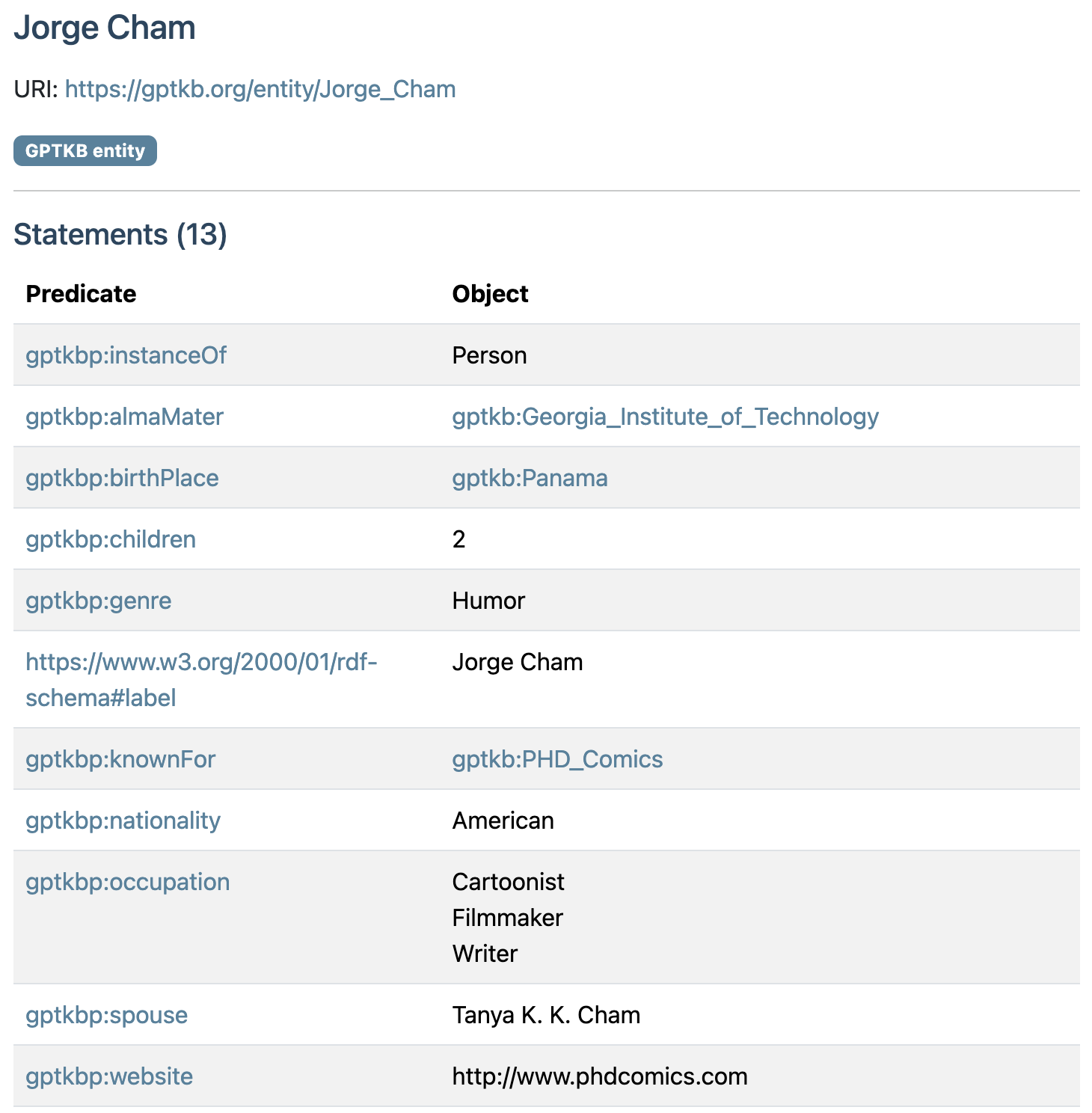}
\caption{Excerpt of \ourkb\ triples for subject \textit{Jorge Cham} of PhDComics.}
\label{tab:jorgecham}
\end{figure}

Figures \ref{tab:vienna} and \ref{tab:jorgecham} provide further examples of \ourkb{} content. In the first case, all triples appear correct, but the predicate \textit{famous\_for} dominates the entity. In second case, most triples are correct, but the spouse is made up. In the online OpenReview appendix to this submission, we provide a larger sample of 10K triples.

\section{Prompts}
\label{app:prompts}

All prompts below are aided by OpenAI's structured output feature, which allows defining a specific JSON schema, that the output has to adhere to. Therefore, the description of the output format in the prompts is less expansive than what is common in many other works.

Figure~\ref{fig:full-prompt} contains the prompt used for triple elicitation. 
Figure~\ref{fig:prompt-ner} shows the prompt for NER. 
Figure~\ref{fig:taxonomy} lists the prompt for taxonomy construction.

\begin{figure*}[ht]
    \centering
    \noindent
    \fbox{\small 
    \begin{minipage}{\textwidth}
        You are a knowledge base construction expert. Given a subject entity, return all facts that you know for the subject as a list of subject, predicate, object triples. The number of facts may be very high, between 50 to 100 or more, for very popular subjects. For less popular subjects, the number of facts can be very low, like 5 or 10.\\
\ \\
Important: \\
- If you don't know the subject, return an empty list. \\
- If the subject is not a named entity, return an empty list.\\
- If the subject is a named entity, include at least one triple where predicate is ``instanceOf''.\\
- Do not get too wordy.\\
- Separate several objects into multiple triples with one object.
        \end{minipage}
    }
    \caption{Prompt for knowledge elicitation.}
    \label{fig:full-prompt}
\end{figure*}

\begin{figure*}[ht]
    \centering
    \noindent
    \fbox{\small 
    \begin{minipage}{\textwidth}
You are an expert on named entity recognition (NER). Your task is to classify if given phrases are named entities (e.g., persons, organizations, works of art), or not (e.g., literals, dates, URLs, verbose phrases). Each phrase is given to you in a line.
        \end{minipage}
    }
    \caption{Prompt for named-entity recognition (NER).}
    \label{fig:prompt-ner}
\end{figure*}

\begin{figure*}[ht]
    \centering
    \noindent
    \fbox{\small 
    \begin{minipage}{\textwidth}
        You are a knowledge base construction expert. 
        Your task is to initialize a seed taxonomy with general categories, which you will update later with given classes.
        Please return only the seed taxonomy in json form with indentation.
    \end{minipage}
    }
    \fbox{\small 
    \begin{minipage}{\textwidth}
        Class: <new class>\\
        \ \\
        You are a knowledge base construction expert.
        Your task is to create a taxonomy for a knowledge base.
        Beforehand, you need to give each given class a score describing how general it is.
        The score is an integer ranging only from 1, for the most general concept, to 10, for the most specific concept.
        Please return only the score of the given class.
    \end{minipage}
    }
    \fbox{\small 
    \begin{minipage}{\textwidth}
        Candidate branches: <branches of current node>\\
        Class: <class to add>\\
        \ \\
        You are a knowledge base construction expert.
        Your task is to integrate a given class into the taxonomy.
        If the given class is a subclass of one of the candidate branches, return only the exact name of that branch.
        Otherwise, return only NULL.  
    \end{minipage}
    }
    \fbox{\small 
    \begin{minipage}{\textwidth}
        Taxonomy: <taxonomy from current node>\\
        Class: <class to add>\\
        \ \\
        You are a knowledge base construction expert.
        Your task is to update the given taxonomy with the given class.
        You can consider the categorization of the taxonomy, but you can not modify the names of the classes in the taxonomy.
        Please return only the updated taxonomy in JSON form.
    \end{minipage}
    }
    \caption{Prompts for seed taxonomy construction (top), class generality scoring (upper middle), recursive class insertion check (lower middle), and (sub)taxonomy update (bottom).}
    \label{fig:taxonomy}
\end{figure*}

\section{Analysis of temporal cutoff}
\label{app:years}

In Figure~\ref{fig:years}, we plot the frequency of values for the \textit{year} property. One can observe two things: 1) A steady increase of frequency with recency, likely mirroring the growth of web corpora over the years, which made more recent content overrepresented. 2) A sudden dropoff between 2023 (549 triples) and 2024 (75 triples), matching the self-declared knowledge cutoff of the GPT-4o-mini LLM. 

\begin{figure}[H]
\centering
\resizebox{\columnwidth}{!}{%
    \begin{tikzpicture}
    \begin{axis}[
        xlabel={Year},
        ylabel={Count},
        ymin=0, ymax=1000,
        xmin=1970, xmax=2030,
        xtick={1970,1980,1990,2000,2010,2020,2030},
        xticklabel style={/pgf/number format/.cd,set thousands separator={}}, 
        ytick={0,250,500,750,1000},
        grid=both,
        width=10cm, height=6cm,
    ]
    \addplot[
        ultra thick,
        color=gg-blue,
    ] coordinates {
    (1970, 154) (1971, 88) (1972, 86) (1973, 63) (1974, 140) (1975, 88) (1976, 84) (1977, 50) (1978, 69) (1979, 68) (1980, 140) (1981, 67) (1982, 64) (1983, 73) (1984, 86) (1985, 92) (1986, 86) (1987, 76) (1988, 75) (1989, 93) (1990, 163) (1991, 107) (1992, 122) (1993, 119) (1994, 140) (1995, 140) (1996, 157) (1997, 157) (1998, 141) (1999, 143) (2000, 324) (2001, 246) (2002, 186) (2003, 181) (2004, 219) (2005, 266) (2006, 293) (2007, 247) (2008, 287) (2009, 268) (2010, 421) (2011, 308) (2012, 381) (2013, 381) (2014, 445) (2015, 549) (2016, 519) (2017, 434) (2018, 593) (2019, 624) (2020, 801) (2021, 917) (2022, 470) (2023, 549) (2024, 75) (2025, 37) (2026, 21) (2027, 22)
    };
    \end{axis}
    \end{tikzpicture}
}
\caption{Number of triples for the \textit{year} property, per year. The sudden drop from 2023 to 2024 matches the self-declared 2023 knowledge cutoff of GPT-4o-mini.}
\label{fig:years}
\end{figure}




\section{Comparison of GPTKB with other KBs}
\label{app:source-kb-sizes}

Table~\ref{tab:KB-sizes} gives a comparison of GPTKB to major KB projects. 
Sources:
\textbf{Wikidata}: See \url{https://www.wikidata.org/wiki/Wikidata:Statistics} and \url{https://grafana.wikimedia.org/d/000000175/wikidata-datamodel-statements}.
\textbf{Wikidata5m}: \cite{wang-etal-2021-kepler}. 
\textbf{Yago4.5}: \cite{yago45}. 
\textbf{DBpedia}: English version as per Table 2 in \cite{dbpedia-statistics}. 
\textbf{NELL}: As per Figure 5 (left) in \cite{nell}. 
\textbf{Reverb}: As per \cite{reverb-linked} and \url{https://web.archive.org/web/20220307185343/https://openie.allenai.org/}.

\end{document}